\documentclass[runningheads]{llncs}

\usepackage{eccv}

\usepackage{eccvabbrv}

\usepackage{graphicx}
\usepackage{booktabs}

\usepackage[accsupp]{axessibility}  

\usepackage{color}
\usepackage{tabularray}
\usepackage[normalem]{ulem}
\usepackage{multirow}
\usepackage{vcell}

\usepackage[misc]{ifsym}

\usepackage{hyperref}

\usepackage{orcidlink}

\begin{document}

\title{Pathformer3D: A 3D Scanpath  Transformer for 360° Images} 


\author{Rong Quan\inst{1} \and
Yantao Lai \inst{1,2}\and
Mengyu Qiu \inst{2}\and
Dong Liang \inst{2,3} \textsuperscript{(\Letter)}}

\authorrunning{R.~Quan et al.}

\institute{College of Artificial Intelligence, Nanjing University of Aeronautics and Astronautics, the Key Laboratory of Brain-Machine Intelligence
Technology, Ministry of Education, Nanjing, 211106, China\\
\email{rongquan0806@gmail.com, yantaolai@nuaa.edu.cn} 
\and MIIT Key Laboratory of Pattern Analysis and Machine Intelligence, College of Computer Science and Technology, Nanjing University of Aeronautics and Astronautics, Nanjing, China\\
\email{\{qmengyu,liangdong\}@nuaa.edu.cn}
\and  Shenzhen Research Institute, Nanjing University of Aeronautics and Astronautics, Shenzhen, China\\
}

\maketitle
\begin{abstract}
  Scanpath prediction in 360° images can help realize rapid rendering and better user interaction in Virtual/Augmented Reality applications. However, existing scanpath prediction models for 360° images execute scanpath prediction on 2D equirectangular projection plane, which always result in big computation error owing to the 2D plane's distortion and coordinate discontinuity. In this work, we perform scanpath prediction for 360° images in 3D spherical coordinate system and proposed a novel 3D scanpath Transformer named Pathformer3D. Specifically, a 3D Transformer encoder is first used to extract 3D contextual feature representation for the 360° image. Then, the contextual feature representation and historical fixation information are input into a Transformer decoder to output current time step's fixation embedding, where the self-attention module is used to imitate the visual working memory mechanism of human visual system and directly model the time dependencies among the fixations. Finally, a 3D Gaussian distribution is learned from each fixation embedding, from which the fixation position can be sampled. Evaluation on four panoramic eye-tracking datasets demonstrates that Pathformer3D outperforms the current state-of-the-art methods. Code is available at \url{https://github.com/lsztzp/Pathformer3D}.
\end{abstract}

\section{Introduction}
\label{sec:intro}

Virtual reality (VR) and augmented reality (AR)~\cite{verma2022advances} technologies have made significant advancements in recent years, providing the users immersive experiences. Meanwhile, understanding and imitating the way human beings explore the 360° images of virtual environments is becoming more and more important, since it can help realize more practical and fast rendering and thus improve user interaction in the immersive environment~\cite{viswanath2009evolution, kay1990user}. Scanpath prediction on 360° images refers to predict human's gaze shift path when exploring 360° images. 

Existing research on scanpath prediction has primarily focused on 2D images~\cite{itti, IRL, VSPT, vqa, ior-roi}, while 360° images possess distinct characteristics in the VR/AR context. 360° images provide immersive interactive environments where users can change their perspectives by physically moving their heads, resulting in a wider distribution of fixations. 360° images encompass richer visual information that requires more time for users to process and absorb. Furthermore, the data structure of 360° images differs from that of 2D images, introducing new requirements for data processing and analysis. Therefore, existing 2D image scanpath prediction methods cannot be directly applied to predict scanpaths in 360° images.

Early methods~\cite{SaltiNet, salient360!, zhu2018prediction} for predicting scanpaths in 360 images involved sampling fixations based on saliency information to obtain the entire scanpath. Subsequently, with the advancement of generative adversarial networks~\cite{aggarwal2021generative}, some researchers~\cite{pathgan, ScanGan360} have utilized generative networks to directly generate the entire path from the 360° image. These methods have shown preliminary results but overlook the modeling of time dependency between fixations, which has long been demonstrated to be a very important characteristic of human visual attention mechanism~\cite{itti}, thereby they often lead to unstable prediction results. Recently, some research works~\cite{fan2017fixation, li2019very, nguyen2018your, ScanDMM} have recognized the need for a comprehensive treatment of the temporal dependencies in viewing behavior, and modeled them through recurrent neural networks (RNNs) or Markov chains. However, these methods integrate all historical fixation information into a single hidden unit, and generate the current fixation only from this hidden unit. In this case, the time dependent relationships among fixations can only be modeled indirectly and the influences from historical fixations will be greatly weakened. Besides, all above methods predict fixations on the 2D equirectangular
projection of 360° image, which has the issue of coordinate discontinuity caused by longitude (where -180° and 180° represent the same fixation despite their significant numerical difference), and different degrees of distortion in different locations. Consequently, the fixation predicted on this 2D equirectangular
projection will have a large margin of error.

To solve the above problems, we propose a novel scanpath prediction model for 360° images named 
Pathformer3D, which predicts fixations in 3D spherical coordinates of the 360° image and directly model the time dependent relationships among fixations, to mimic human viewing behavior in immersive environment more realistically. 
Specifically, given a 2D equirectangular
projection of 360° image, we first transform it into 3D spherical coordinate system, and then exploit a spherical convolution to extract its visual features. After extracting the 3D visual features, we exploit a 3D Transformer encoder to learn each 360° image region a contextual feature representation, by considering their long range spatial dependencies between each other. 
Next, we exploit a Transformer Decoder to learn each time step's fixation embedding directly from the whole image's visual feature and the historical fixations, where the self-attention module is used to mimic human visual working memory mechanism, which can model the time dependencies between the current fixation and all  historical fixations. 
After obtaining each time step's fixation embedding, we employ a 3D mixture density network to learn the 3D Gaussian distribution for each fixation, from which the fixation location is sampled. Using 3D Gaussian distribution to model the positional likelihood of fixations in 3D space takes into account the differences in scanpaths of different individuals, and can thus result in more robust and real scanpaths. 

In conclusion, this work has the following contributions: 
\begin{itemize}
\item We perform visual scanpath prediction for 360° images in a 3D spherical coordinate system for the first time, which can realize a more real imitation of human visual exploration process in immersive environment and simultaneously avoid the error caused by the distortion and positional discontinuity in 2D equirectangular projection of 360° image.
\item We propose a novel scanpath prediction method named Pathformer3D, to learn contextual feature representation of the 360° images in a 3D spherical coordinate system and directly consider the influences from historical fixations on current fixation, which is more in line with the visual working memory mechanism of human vision system. 
\item The proposed method is comprehensively evaluated on four eye-tracking datasets of 360° images, consistently achieving state-of-the-art performance, demonstrating the effectiveness of our approach.
\end{itemize}

\section{Related Works}

\subsubsection{Scanpath Prediction for 2D Images}
In recent years, numerous models have been proposed for predicting scanpaths in 2D images. According to the different sources of inspiration, they can be categorized into different classes.

Some of these models draw inspiration from neuroscience and vision science, often leveraging human gaze behavior to guide scanpath prediction.
For instance, some researchers\cite{itti,scenewalk,ior-roi} have utilized Inhibit-of-Return(IOR) strategies to guide the generation of scanpaths.
Addtionally, Wang \etal~\cite{TDE} compute response maps of sparse coding filters applied to foveated images at each step and select new fixations from the residual perceptual information map(RPI) using the information maximization principle. 
Adeli \etal~\cite{adeli} first apply a retina transformation to input images, then compute corresponding priority maps and select the next fixation. Furthermore, some researchers\cite{latest,g-eymo} attempt to generate new fixations by leveraging low-level semantic information.

Another subset of work is inspired by the statistical distribution characteristics of human scanpaths. Many researchers\cite{bro,cle,le15,le16,IRL,liu,VSPT} in this group attempt to use image saliency information to predict new scanpaths. For example, VSPT\cite{VSPT} uses SalGAN\cite{salgan} to extract saliency information and employs an autoregressive approach\cite{autoregressive} to generate fixations. 
Others in this category adopt alternative methods, such as Sun \etal~\cite{sun}, who conduct super-Gaussian quality analysis on image patches and select the $k$th fixation to be at the patch with the highest response to the $k$th super Gaussian component. Coutrot \etal~\cite{coutrot} model scanpaths using image and task dependent hidden markov models(HMM). Clarke \etal~\cite{clarke} implement a Gaussian jump distribution, which is polynomially dependent on the previous fixed distribution.

In addition to the aforementioned models, there are also some deep learning models\cite{DeepGazeIII,pathgan,yang} that have achieved relatively good performance, despite their limited interpretability.

\subsubsection{Scanpath Prediction for 360° Images}
Unlike 2D images, there are currently limited models specifically designed for scanpath prediction in 360° images. However, with the gradual development of VR/AR technologies, this field has attracted increasing attention. In this domain, SaltiNet\cite{SaltiNet} is the first proposed method, which uses saliency volume to process 360° images. PathGAN\cite{pathgan} adopts the method of Generative Adversarial\cite{GAN} to generate scanpath of 360° image. Zhu \etal~\cite{Zhu} proposed a new method to estimate the saliency of 360° images, and introduced mechanisms such as visual uncertainty and visual balance to predict the scanpath. The subsequent ScanGAN\cite{ScanGan360} is also a method that uses Generative Adversarial Networks\cite{GAN} to generate scapath. Different from PathGAN\cite{pathgan}, ScanGAN\cite{ScanGan360} adds a soft-dtw loss function\cite{SoftDTW} to the generator to generate scanpath that are more similar to real scanpath. ScanDMM\cite{ScanDMM} is a method that uses Deep Markov model\cite{DMM} to generate scanpath, which takes into account the past historical information.

All above methods predict fixations as well as scanpaths in 2D equirectangular
projection of 360° image, which has the weaknesses of distortion and coordinate discontinuity and always introduce big prediction error. Different from them, we perform scanpath prediction for 360° images in 3D spherical coordinate system, which is more consistent with the real-world scenarios of human exploring immersive environments. 

\section{Approch}

\begin{figure}[tb]
  \centering
  \includegraphics[width=1\textwidth]{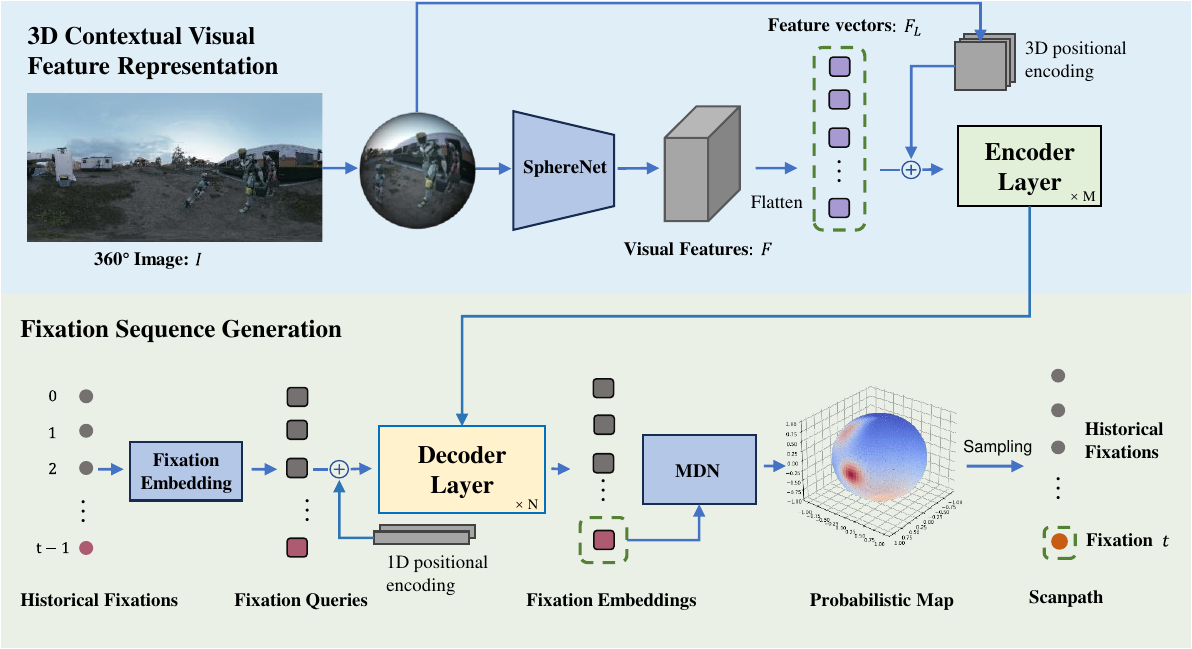}
  \caption{Overall architecture of our methods. The Transformer Encoder is utilized to contextualize the feature extracted by the SphereNet\cite{S-CNN}. The contextual feature representation and the historical fixations are fed into the Transformer Decoder to output fixation embeddings. Each fixation embedding is input into a 3D Mixture Density Network to output its 3D Gaussian distribution, from which the fixation is sampled.
}
  \label{architecture}
\end{figure}

\subsection{Problem Definition}
A scanpath can be represented as a sequence of fixations $X = (p_{1},p_{2},\-\ldots,p_{T})\in\mathbb{R}^{T\times 3}$, where $T$ represents the length of the scanpath. For a given 360° image, the task of a scanpath prediction model is to generate a scanpath $\widehat{X}$ that resembles a real human scanpath. We use a 3D Cartesian coordinate system to represent coordinates, where each fixation is given in the form of $p=(x, y, z)$, with x, y, and z being real numbers ranging from -1 to 1. This representation form helps address the issue of coordinate discontinuity.

\subsection{Overview}
The architecture of our model is shown in \cref{architecture}. For each input 360° image, we first transform it into 3D spherical coordinate system, and then extract the image features using SphereNet\cite{S-CNN}. Subsequently, we feed the extracted feature representation into a 3D Transformer Encoder to contextualize them. The Transformer Decoder takes historical fixation information and contextual feature representations as input and generates embedding information for new fixations in an autoregressive manner\cite{autoregressive}. Then, a 3D Mixture Density Network (MDN) takes the embedding information of the fixations as input and outputs their 3D Gaussian position distributions. Finally, the fixations are sampled from the 3D Gaussian distributions. Next, we will introduce a detailed module-wise description of our model.

\subsection{3D Contextual Visual Feature Representation}

\subsubsection{3D feature extraction.}
We perform scanpath prediction in 3D spherical coordinate system. Therefore, after projecting the 360° image into 3D spherical coordinate system, we utilize SphereNet\cite{S-CNN} to extract its 3D feature representation. Specifically, for a 360° image $I$, after inputting it into SphereNet, we extract 3 different scales of feature maps and concatenate them along the channel axis,  resulting in a feature representation $F$ with dimensions of $C=448,H=128,W=256$.

Before inputting into the 3D Transformer Encoder, we first execute a simple size transformation to $F$.
Firstly, we execute average pooling to $F$ in a convolutional kernel of size $8\times8$, to reduce the size of the feature maps to  $C = 448, H=16,W=32$. Next, we flatten the pooled feature maps, resulting in a feature $F_{L}$ with size of $L = 16\times32 = 512, C = 448$. This series of operations can be formulated as:
\begin{align}
F &= SphereNet(I, \theta ) \; \\
F_{L}&=\text{Flatten}(\text{AvgPool}(F))
\end{align}
here, $\theta$ represents the parameters of SphereNet.

\subsubsection{3D feature encoder.} A linear embedding layer is first applied to map $F_{L}$ into feature embeddings. 
Subsequently, the obtained feature embeddings are added to a 3D positional encoding and fed into $M=4$ standard Transformer Encoder Layers to extract contextual feature representations. The above process can be expressed by the following formulas.
\begin{align}
F_{L}^{\prime}&=\text{Embedding}(F_{L}) \; \\
P_{0}&={F_{L}^{\prime}}+{E_{pos\_3d}}\hfill & &E_{pos\_3d}\in{R}^{L\times{D}} \; \\
P_{i}&=\text{EncoderLayer}(P_{i-1})\hfill & &i=1 \ldots M
\end{align}
where $D$ represents the feature embedding dimension of the 3D Transformer Encoder, $M$ represents the number of EncoderLayer in the architecture, and $P_{i}$ represents the output of the i-th EncoderLayer.

\subsection{Fixation Sequence Generation}
\subsubsection{Fixation decoder.} The contextual feature representation and historical fixation information are input into a Transformer decoder to output the fixation embedding. The fixation embedding is then fed into a 3D mixture density network to output the fixation's 3D Gaussian distribution, from which the fixation is sampled. 
We use autoregressive method\cite{autoregressive} to predict one fixation each time and iteratively generate the whole fixation sequence. 

On time step $t$, we initialize the fixation query vector $q_t$ using fixation $p_{t-1}$ of the previous time step. Specially, for the first fixation query $q_1$, we initialize it using the center of 3D space, \textit{i.e.}, $p_0=(0, 0, 0)$. Subsequently, the initialized query vector $q_t$ is concatenated with the historical query vectors to generate the joint query vector $Q_{0}^{\prime}$. After adding one-dimensional positional encoding information, $Q_{0}^{\prime}$ becomes $Q_{0}$. Finally, after passing through $N=4$ standard Transformer Decoder Layers, we obtain the hidden states $Q_{N}=(Z_{1},Z_{2},\ldots,Z_{t})$ of the fixations. This process is defined as follows:
\begin{align}
q_{t} &=\text{Embedding}(p_{t-1})& & t\in[1,2,\ldots,T] \; \\
Q_{0}^{\prime} &=[q_{1}; q_{2}; \cdots ; q_{t}] \; \\
Q_{0} &=Q_{0}^{\prime}+{E_{pos\_1d}}& & E_{pos\_1d}\in{R}^{t\times{D}} \; \\
Q_{i} &=\text{DecoderLayer}(Q_{i-1}, P_{M})& & i=1 \ldots N
\end{align}
where $P_{M}$ represents the output of the Transformer Encoder, $T$ represents the length of the predicted fixation sequence, $p_{t-1}$ represents the fixation at the time step $t-1$, $N$ represents the number of decoder layers. $Q_{i}$ represents the output of the i-th DecoderLayer.

The fixation decoder process in the training process has some difference from that in the generation process. During the training phase, we input the fixations $(p_{0},p_{1},\ldots,p_{T-1})\in R^{T\times 3}$ into the Transformer Decoder, and it can predict the hidden states $Q_{N}=(Z_{1},Z_{2},\ldots,Z_{T})$ of the true scanpath $(p_{1},p_{2},\ldots,p_{T})\in R^{T\times 3}$ in a parallel manner, which is a parallel process. During the generation phase, we iteratively generate the hidden states of new coordinates $Z_{1},Z_{2},... Z_{T}$ in sequence starting from the initial coordinate $p_{0}$, which is a sequential process.

\subsubsection{Fixation generation.}
After obtaining $Z_t$, we then employ a 3D mixture density network to model the probability distribution of the fixations. The 3D mixture density network decodes the hidden state $Z_t$ of the fixation at time step $t$ into $K$ sets of Gaussian kernel parameters. We use $\mu_{t}^{i} = (\mu_{x,t}^{i}, \mu_{y,t}^{i}, \mu_{z,t}^{i})$, $\Sigma_{t}^{i}$, and $\pi_{t}^{i}$ to represent the mean vector, covariance matrix, and weight of the $i$-th Gaussian kernel at time step $t$, respectively.
These three parameters can be obtained through three separate MLP layers, which can be defined as:
\begin{align}
\left\{\mu_{t}^{i}, \Sigma_{t}^{i}, \pi_{t}^{i}\right\} &= f_{mdn}\left(Z_{t} ; \theta_{mdn}\right) & & i=1 \ldots K
\end{align}
where $\theta_{mdn}$ refers to the parameters of the mixture density network.

Next, the probability density function for a 3D Gaussian distribution $P(p_{t})$ can be defined as follows:
\begin{align}
\mathcal{N}(p_{t} \mid \mu_{t}^{i}, \Sigma_{t}^{i}) &= \frac{\exp\left(-\frac{1}{2}(p_{t}-\mu_{t}^{i})^\top (\Sigma_{t}^{i})^{-1} (p_{t}-\mu_{t}^{i})\right)}{\sqrt{(2\pi)^3 |\Sigma_{t}^{i}|}} \; \\
P(p_{t})&=\sum_{i=1}^{K}\pi_{t}^{i}\mathcal{N}(p_{t} \mid \mu_{t}^{i}, \Sigma_{t}^{i})
\end{align}
where the $|\Sigma_{t}^{i}|$ means the determinant of the covariance matrix $\Sigma_{t}^{i}$ and the $\mathcal{N}(p_{t} \mid \mu_{t}^{i}, \Sigma_{t}^{i})$ represents the Gaussian distribution function associated with the $i$-th 3D Gaussian function.

The prediction fixation $\hat{p}_{t}$  is sampled probabilistically from a 3D surface, which can be expressed as:
\begin{align}
\hat{p}_{t} \sim P(p_{t})
\end{align}
here, it is worth noting that we do not simply sample the fixation with the highest probability, since such sampling strategy always make the model rapidly converge to the same fixation positions for all images. In our study, we generated a total of 32,768 fixation points along the latitude-longitude grid, with a resolution of $128\times256$. We independently computed the probability for each fixation point. Finally, we randomly selected a fixation point based on its probability, with higher probabilities being more likely to be chosen.

\subsection{Loss Function}
During training, the objective is to minimize the discrepancy between the predicted distribution and the real human scanpath. To achieve this, we employ the negative log-likelihood loss function, which encourages higher probabilities for the ground truth fixations. 
The loss function can be defined as follows:
\begin{align}
\mathcal{L} &= -\frac{1}{T} \sum_{t=1}^{T} \log \left( \sum_{i=1}^{K} \pi_{t}^{i} \mathcal{N}\left(p_{t}^{*} | \mu_{t}^{i}, \Sigma_{t}^{i}\right) \right)
\end{align}
where $p_{t}^{*}$ represents the real human fixation position of time step $t$.

\section{Experiments}

\subsection{Datasets}
We conducted experiments on four 360° image datasets: Sitzmann~\cite{Sitzmann}, Salient360!~\cite{salient360!}, AOI~\cite{AOI}, and JUFE~\cite{JUFE}. 
The Sitzmann dataset\cite{Sitzmann} comprises 22 images with a total of 1920 scanpaths. Only the training set of the Salient360! dataset\cite{salient360!} is available, which contains 85 images and 3036 scanpaths. The AOI dataset\cite{AOI} consists of 600 high-resolution images and approximately 18,000 scanpaths. The JUFE dataset\cite{JUFE} includes 1032 images and 30,960 scanpaths. For training, 19 images from Sitzmann\cite{Sitzmann} and 60 images from Salient360!\cite{salient360!} were utilized, and all remaining data were used for validation. To increase the number of training sets, we rotated the image along with the corresponding scanpaths six times along the direction of longitude. We followed ScanDMM\cite{ScanDMM} to pre-process the gaze data of Sitzmann, Salient360!, and JUFE datasets, and obtain visual scanpaths containing 30, 25, and 15 fixations for Sitzmann, Salient360!, and JUFE.
The AOI dataset has already been pre-processed, with an average scanpath length of approximately 8.

\subsection{Evaluation Metrics}
We employed six metrics to assess the performances of our model and other models, including Levenshtein Distance (LEV)\cite{LEV}, Dynamic Time Warping (DTW)\cite{DTW}, Time Delay Embedding (TDE), ScanMatch\cite{ScanMatch}, Recursive Metric (REC)\cite{REC} and Sequence Score(SS)\cite{yang}. 
LEV\cite{LEV} is a metric that quantifies the dissimilarity between two strings by measuring the minimum number of single-character edit operations needed to transform one string into another. ScanMatch\cite{ScanMatch} involves converting eye-tracking scanpaths into string sequences and then employing an algorithm similar to the Levenshtein distance to compare the similarity of the two sequences. DTW\cite{DTW} and TDE\cite{TDE} are metrics based on time series analysis, while REC\cite{REC} is a metric for recursive quantification analysis. 
SS\cite{yang} is calculated by converting scanpaths into strings of fixation cluster IDs, and then measuring the similarity between two strings using a string matching algorithm\cite{sma}. 

For each image, every model generates $m=10$ predicted scanpaths $\widehat{X}$, which are evaluated against all real human scanpaths $X$, and then averaged to determine the final prediction score. Note that our model predicts scanpath with a length of 30. When comparing across different datasets, we truncate the predicted scanpath to the length of the actual scanpath before conducting the comparison. Taking DTW\cite{DTW} as an example, the score can be calculated as follows:
\begin{equation}
DTW=\frac{1}{n}\frac{1}{m}\sum_{i=1}^{n}\sum_{j=1}^{m}DTW(\widehat{X}_{j},X_{i})\label{metric}
\end{equation}
where $\widehat{X}_{j}$ is the $j$-th predicted scanpath, $X_{i}$ is the $i$-th real human scanpath, $n$ is the number of the real human scanpaths, $m$ is the number of predicted scanpaths.

\subsection{Implementation Details}
The internal vector dimension of 3D Transformer Encoder is set to $D=128$. We utilize a multi-head attention mechanism with 8 attention heads. The hidden layer size of the feed-forward network (FFN) is set to 64.
In the 3D mixture density network model, we configure $K=5$ Gaussian kernels, and the internal hidden layer dimension is set to 16.
During the training process, our batch size is set to 18. As for the optimizer, we use AdamW\cite{adam} with a learning rate of 1e-5. We employ a learning rate warm-up and stage-wise adjustment strategy. The warm-up period lasts for 10 epochs, and the learning rate is halved every 10 epochs. The total training duration is 50 epochs.

Our experiments were conducted on a single NVIDIA GeForce RTX 3090, and the development was conducted using the PyTorch framework.

\subsection{Performance Comparision}
We compared our model with SaltiNet\cite{SaltiNet}, ScanGAN\cite{ScanGan360}, and ScanDMM\cite{ScanDMM}, utilizing the models provided officially by their respective sources to obtain prediction results. These three models are all designed for scanpath prediction on 360° images. PathGAN\cite{pathgan} and Zhu \etal~\cite{Zhu} are two additional models specifically designed for generating scanpath on 360° images. However, it has been demonstrated that these models are unable to effectively generate realistic scanpaths\cite{ScanGan360}. Therefore, we have not included them in our comparison.

We additionally established two baseline reference values to assess the relative quality of the generated scanpath. 
The first one is the random baseline (Random walk). This baseline randomly generate ten scanpaths for each image, which provides the reference for lowest performance. 
The second one is the human baseline (Human), which compares each real human scanpath in the dataset with all the other real human scanpaths and taking the average. 
This baseline represents the average similarity among real human scanpaths, serving as a reference for highest performance.

\subsubsection{Quantitative evaluation.}
\cref{quantitative} shows the quantitative comparison results between our method and other competitors on the metrics of LEV\cite{LEV}, DTW\cite{DTW}, TDE\cite{TDE}, ScanMatch\cite{ScanMatch}, REC\cite{REC} and SS\cite{yang}. As we can see, our method outperforms all the competitors on all four datasets, achieving performance close to `Human'. 

\begin{table}
\centering
\caption{Performance of scanpath prediction models on different datasets. The best result for each metric is highlighted in bold, while the second-best result is underlined.}
\label{quantitative}
\renewcommand{\arraystretch}{1.25}
\resizebox{\textwidth}{!}{
\begin{tabular}{ccccccc|cccccc} 
\hline
\multirow{2}{*}{Method} & \multicolumn{6}{c|}{Sitzmann}                                                                            & \multicolumn{6}{c}{Salient360!}                                                                           \\ 
\cline{2-13}
                        & LEV ↓           & DTW ↓             & TDE ↓           & ScanMatch ↑    & REC ↑          & SS ↑           & LEV ↓           & DTW ↓             & TDE ↓           & ScanMatch ↑    & REC ↑          & SS ↑            \\ 
\hline
Random walk             & 51.522          & 2372.387          & 28.038          & 0.395          & 1.514          & 0.198          & 43.670          & 2062.080          & 29.753          & 0.375          & 1.381          & 0.176           \\ 
\hline
SaltiNet\cite{SaltiNet}                & 51.151          & 2246.057          & 26.303          & 0.374          & 1.734          & 0.198          & 41.042          & 1860.397          & 25.001          & 0.429          & 2.250          & 0.220           \\
ScanGAN\cite{ScanGan360}                & 45.852          & 1950.488          & \textbf{19.060} & 0.477          & 2.960          & 0.265          & 38.977          & 1752.371          & \uline{19.527}  & 0.460          & 3.041          & 0.244           \\
ScanDMM\cite{ScanDMM}                 & \uline{45.748}  & \uline{1939.980}  & \uline{19.098}  & \uline{0.485}  & \uline{3.156}  & \uline{0.270}  & \uline{38.545}  & \uline{1706.661}  & 20.238          & \uline{0.467}  & \uline{3.125}  & \uline{0.257}   \\
Ours                    & \textbf{44.691} & \textbf{1939.662} & 19.887          & \textbf{0.486} & \textbf{4.154} & \textbf{0.280} & \textbf{36.277} & \textbf{1515.883} & \textbf{19.261} & \textbf{0.484} & \textbf{4.250} & \textbf{0.302}  \\ 
\hline
Human                   & 42.006          & 1841.534          & 16.192          & 0.526          & 5.163          & 0.366          & 35.090          & 1376.439          & 16.650          & 0.513          & 5.066          & 0.377           \\ 
\hline
\multirow{2}{*}{Method} & \multicolumn{6}{c}{AOI}                                                                                  & \multicolumn{6}{c}{JUFE}                                                                                  \\ 
\cline{2-13}
                        & LEV ↓           & DTW ↓             & TDE ↓           & ScanMatch ↑    & REC ↑          & SS ↑           & LEV ↓           & DTW ↓             & TDE ↓           & ScanMatch ↑    & REC ↑          & SS ↑            \\ 
\hline
Random walk             & 15.866          & 666.652           & 38.164          & 0.293          & 5.650          & 0.241          & 26.439          & 1270.893          & 38.164          & 0.352          & 1.847          & 0.191           \\ 
\hline
SaltiNet\cite{SaltiNet}                & 14.978          & 622.079           & 35.724          & 0.337          & 6.306          & 0.245          & 26.115          & 1264.492          & 32.261          & 0.364          & 1.972          & 0.192           \\
ScanGAN\cite{ScanGan360}                 & 14.473          & 569.549           & 32.602          & 0.361          & \uline{7.379}  & \textbf{0.258} & 24.259          & 1110.826          & 25.937          & 0.431          & 3.208          & 0.217           \\
ScanDMM\cite{ScanDMM}                 & \uline{14.185}  & \uline{569.170}   & \uline{32.558}  & \uline{0.371}  & 7.328          & \uline{0.249}  & \uline{23.372}  & \uline{1081.465}  & \uline{25.787}  & \uline{0.452}  & \uline{3.527}  & \textbf{0.221}  \\
Ours                    & \textbf{13.904} & \textbf{561.895}  & \textbf{30.987} & \textbf{0.376} & \textbf{8.286} & 0.225          & \textbf{22.595} & \textbf{1054.223} & \textbf{22.955} & \textbf{0.462} & \textbf{4.449} & \uline{0.218}   \\ 
\hline
Human                   & 14.317          & 501.952           & 30.248          & 0.367          & 13.008         & 0.377          & 18.445          & 1030.970          & 20.305          & 0.564          & 7.864          & 0.330           \\
\hline
\end{tabular}}
\end{table}

\subsubsection{Qualitative evaluation.}
\begin{figure}[tb]
  \centering
  \includegraphics[width=1\textwidth]{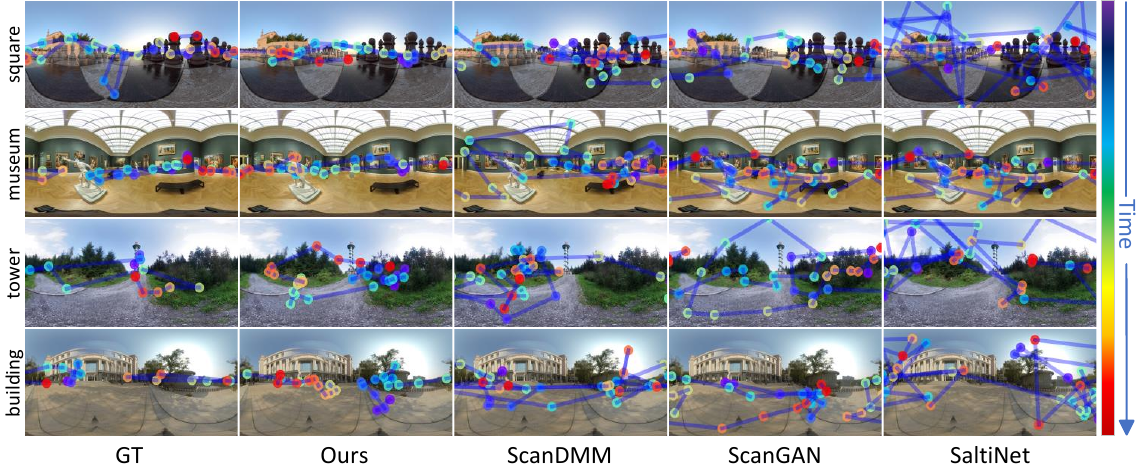}
  \caption{Qualitative representation results of different models on four datasets. From top to bottom, the four images are sourced from Sitzmann\cite{Sitzmann}, Salient360!\cite{salient360!}, AOI\cite{AOI}, and JUFE\cite{JUFE}, respectively. From left to right, we display the a scanpath sampled from ground truth, our model, ScanDMM\cite{ScanDMM}, ScanGAN\cite{ScanGan360}, and SaltiNet\cite{SaltiNet}.
  }
  \label{qualitatibe}
\end{figure}

We visualized the prediction results of all methods on one image of each dataset, as shown in \cref{qualitatibe}. Similarly, we visualized the ground truth of human scanpath accordingly. The transition from purple to red in the scanpath indicates the beginning and end of the scanpath, respectively. It is noticeable that most ground truth scanpaths are located within the salient regions near the equator. SaltiNet\cite{SaltiNet}, although designed for 360° images, suffers from large displacements. In comparison, the methods proposed by ScanGAN\cite{ScanGan360} and ScanDMM\cite{ScanDMM} have the capability to generate relatively realistic scanpaths to some extent. However, our model, in contrast to these approaches, excels at capturing salient information within 360° images and is capable of generating scanpath that are closer to the real human scanpaths.
In addition, a notable flaw in the tower image is that our model tends to focus more on the prominent plants at both ends rather than the central tower, which is a common issue shared by other models as well.

\subsection{Saliency Comparison}
A good scanpath prediction model often assigns more attention to salient regions during the prediction process. In our study, we generated 1000 scanpath and performed post-processing on all fixations to generate corresponding saliency maps\cite{gauss}. The fixation maps was obtained from all unprocessed fixations. We used sizes H=128 and W=256 for saliency maps (salmap) and fixation maps (fixmap). We compared our results with ScanGAN\cite{ScanGan360} and ScanDMM\cite{ScanDMM} on the Salient360!\cite{salient360!} and JUFE\cite{JUFE} datasets, under the saliency metrics AUC\_Judd\cite{AUC_SIM}, NSS\cite{NSS}, CC\cite{CC}, SIM\cite{AUC_SIM},and KLD\cite{KLD}. 

\begin{table}
\centering
\caption{Quantitative comparison results of significance comparison on Salient360!\cite{salient360!} and JUFE\cite{JUFE} datasets. The best result for each metric is highlighted in bold, while the second-best result is underlined.}
\label{saliency_compare}
\resizebox{0.8\textwidth}{!}{
\begin{tabular}{llccccc} 
\toprule
Dataset                      & Method  & AUC\_Judd ↑          & NSS ↑          & CC ↑           & SIM~↑          & KLD ↓            \\ 
\hline
\multirow{3}{*}{Salient360!\cite{salient360!}} & ScanGAN\cite{ScanGan360} & 0.797          & 0.986          & 0.260          & 0.173          & 21.210           \\
                             & ScanDMM\cite{ScanDMM} & \uline{0.830}  & \textbf{1.210} & \uline{0.322}  & \uline{0.183}  & \uline{20.882}   \\
                             & Ours    & \textbf{0.838} & \uline{1.195}  & \textbf{0.352} & \textbf{0.259} & \textbf{18.985}  \\ 
\hline
\multirow{3}{*}{JUFE\cite{JUFE}}        & ScanGAN\cite{ScanGan360} & 0.795          & 0.837          & 0.198          & 0.236          & 19.529           \\
                             & ScanDMM\cite{ScanDMM} & \uline{0.866}  & \textbf{1.203} & \textbf{0.275} & \uline{0.286}  & \uline{18.248}   \\
                             & Ours    & \textbf{0.872} & \uline{1.035}  & \uline{0.245}  & \textbf{0.320} & \textbf{16.339}  \\
\bottomrule
\end{tabular}}
\end{table}

\begin{figure}[tb]
  \centering
  \includegraphics[width=1\textwidth]{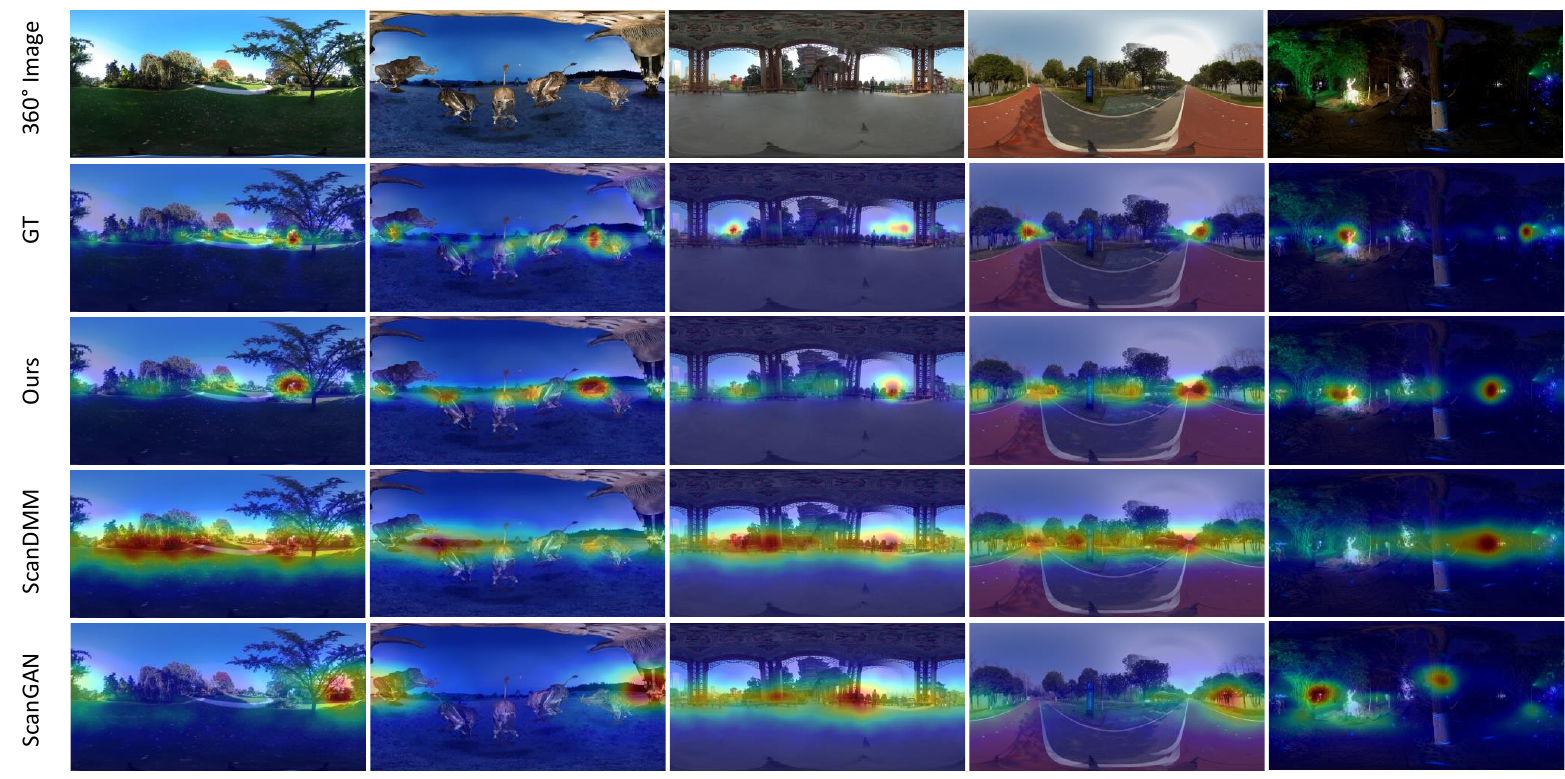}
  \caption{Qualitative comparison results of significance comparison.The first two images is from Salient360!\cite{salient360!} and the next three images are from JUFE\cite{JUFE}. From top to bottom are the real image, the significance maps of the real images, Ours, ScanDMM\cite{ScanDMM}, ScanGAN\cite{ScanGan360}.
  }
  \label{saliency_qualitiative}
\end{figure}
\begin{table}
\centering
\caption{Efficiency comparison in terms of model size and running time}
\label{efficiency}
\resizebox{0.8\textwidth}{!}{
\begin{tabular}{lllll} 
\toprule
                 & SaltiNet~\cite{SaltiNet} & ScanGAN~\cite{ScanGan360} & ScanDMM~\cite{ScanDMM} & Ours     \\ 
\hline
Paramters(MB)    & 98.73    & 15.134  & 17.779  & 19.280   \\
Running Time(ms) & 12720    & 6.482   & 118.250 & 266.420  \\
\bottomrule
\end{tabular}}
\end{table}
Based on \cref{saliency_compare}, our model has demonstrated favorable performance across the majority of metrics. Our model did not achieve optimal results in a few metrics, one potential explanation is that our model exhibits a stronger emphasis on larger salient objects, which may lead to comparatively less attention towards smaller targets compared to models with more evenly distributed saliency maps. As evident in \cref{saliency_qualitiative}, ScanDMM\cite{ScanDMM} and ScanGAN\cite{ScanGan360} exhibit a characteristic of having more dispersed saliency maps. Furthermore, it can be observed that our model has the ability to focus more on salient regions compared to the other models. As a result, the post-processed saliency maps obtained from our model exhibit a closer resemblance to human-generated saliency maps.

\subsection{Efficiency Comparison}
We also compared our model's size and running time for generating a scanpath with other methods. In \cref{efficiency}, we present the model size and the runtime for generating a single scanpath (measured in milliseconds). As can be seen, we have a similar model size with other models. In terms of runtime, our model is capable of generating approximately four scanpaths of length 30 per second, which is a little slower compared to ScanGAN and ScanDMM. This is primarily because our model generates fixations in an autoregressive manner\cite{autoregressive}. Additionally, the use of a mixture density network is another factor leading to slower running speed. However, compared to a single Gaussian distribution, multiple Gaussian distributions can more accurately represent the distribution of fixations.

\subsection{Ablation Studies}
We conducted ablation experiments on the Salient360!\cite{salient360!} and JUFE\cite{JUFE} datasets. The experimental results are presented in \cref{Ablation_table}, showing the optimal results obtained for each configuration. Next, we will provide a brief overview of each model architecture. For more detailed architecture information, please refer to the supplementary file. 

\begin{table}
\centering
\caption{Ablation studies on Salient360!\cite{salient360!} and JUFE\cite{JUFE} datasets. The best result for each metric is highlighted in bold, while the second-best result is underlined.}
\label{Ablation_table}
\renewcommand{\arraystretch}{1.25}
\resizebox{\textwidth}{!}{
\begin{tabular}{ccccccc|cccccc} 
\hline
\multirow{2}{*}{Method} & \multicolumn{6}{c|}{Salient360!}                                                                                                                                                                    & \multicolumn{6}{c}{JUFE}                                                                                  \\ 
\cline{2-13}
                        & LEV ↓           & DTW ↓             & TDE ↓           & ScanMatch ↑    & REC ↑                                                                                                     & ~SS ↑          & LEV ↓           & DTW ↓             & TDE ↓           & ScanMatch ↑    & REC ↑          & SS ↑            \\ 
\hline
Pure ViT                & 37.643          & 1739.714          & \uline{19.291}  & 0.470          & 3.710                                                                                                     & 0.267          & 22.891          & 1072.753          & 23.751          & 0.457          & 3.847          & \textbf{0.221}  \\
Pure 2D CNN             & 36.848          & 1727.241          & 19.641          & 0.477          & 4.119                                                                                                     & 0.272          & 22.613          & 1060.205          & \uline{23.096}  & \textbf{0.463} & 4.181          & \uline{0.220}   \\
Saliency~ ~~            & 38.196          & 1864.914          & 20.059          & 0.475          & 3.156                                                                                                     & 0.253          & 23.833          & 1094.290          & 24.636          & 0.442          & 3.669          & 0.208           \\ 
\hline
w/o EncoderLayer        & 37.352          & 1798.340          & 20.727          & 0.463          & \uline{4.171}                                                                                             & 0.262          & 22.926          & 1072.263          & 23.196          & 0.455          & 4.260          & 0.218           \\ 
\hline
w/o MDN + MSE Loss      & 39.463          & 1793.948          & 25.914          & 0.426          & 3.781                                                                                                     & 0.095          & 23.915          & 1132.127          & 25.675          & 0.416          & 3.833          & 0.102           \\ 
\hline
K=1                     & 38.643          & 1804.074          & 20.074          & 0.462          & 3.326                                                                                                     & 0.248          & 23.239          & 1084.734          & 23.577          & 0.451          & 3.805          & 0.218           \\
K=3                     & \textbf{36.250} & 2073.905          & 22.326          & \textbf{0.490} & 4.039                                                                                                     & 0.246          & 22.758          & 1077.365          & 23.302          & 0.460          & \uline{4.360}  & 0.214           \\
K=8                     & 38.117          & 1700.877          & 20.455          & 0.467          & 3.389                                                                                                     & 0.258          & 22.917          & 1065.240          & 23.470          & 0.455          & 3.914          & 0.210           \\ 
\hline
Parallel                & 36.722          & \uline{1577.993}  & 23.726          & \uline{0.489}  & 3.939                                                                                                     & \uline{0.288}  & \textbf{22.567} & \uline{1054.644}  & 28.775          & 0.451          & 3.688          & \textbf{0.221}  \\ 
\hline
Ours                    & \uline{36.277 } & \textbf{1515.883} & \textbf{19.261} & 0.484          & \textcolor[rgb]{0.275,0.322,0.396}{\textcolor{black}{\textbf{4.253}}}\textcolor[rgb]{0.275,0.322,0.396}{} & \textbf{0.302} & \uline{22.595}  & \textbf{1054.223} & \textbf{22.955} & \uline{0.462}  & \textbf{4.449} & 0.218           \\
\hline
\end{tabular}}
\end{table}

\subsubsection{The feature extractor.}
To demonstrate the effectiveness of our 3D feature extractor, SphereNet, we try to employ three distinct architectures for feature extraction. Firstly, we extracted features using the patch-based approach of the Vision Transformer (ViT) method\cite{ViT} (denoted as `Pure ViT'). Secondly, we replaced SphereNet with a standard 2D convolutional network (denoted as `Pure 2D CNN').
Lastly, We followed VSPT\cite{VSPT} to extract image features based on the saliency information. Specifically, we employed EPSNet\cite{epsnet} to extract saliency information (denoted as `Saliency'). As shown in \cref{Ablation_table}, `Ours' obtains better performance than `Pure ViT', `Pure 2D CNN', and `Saliency', which demonstrates the superiority of our feature extractor.

\subsubsection{The 3D Transformer encoder.}
To validate the effectiveness of our 3D Transformer Encoder, we attempted to remove the EncoderLayer in the 3D Transformer Encoder and directly utilize the feature embedding of the Transformer Encoder as input of the Transformer Decoder (denoted as `w/o EncoderLayer'). As we can see from \cref{Ablation_table}, without our 3D Transformer Encoder, the scanpath prediction performance dropped significantly, which demonstrates the effectiveness of our 3D Transformer Encoder.

\subsubsection{3D MDN.}
To validate the effectiveness of the 3D MDN, we directly exploited linear regression  to predict the fixations after the Transformer Decoder. The Mean Squared Error (MSE) was utilized as the loss function (denoted as `w/o MDN + MSE Loss'). As can be seen in \cref{Ablation_table}, `Ours' outperforms `w/o MDN + MSE Loss' a lot, proving the advantage of our 3D MDN. 

\subsubsection{Different number of Gaussian kernels.}
We compared with the model results obtained by using different numbers of Gaussian kernels in the mixture density network. These comparisons are respectively presented in the rows labeled "K=1,3,8," while our model has 5 Gaussian kernels. As we can see from \cref{Ablation_table}, using 5 Gaussian kernels can provide a better representation of the fixations' distribution.

\subsubsection{Parallel in the Transformer decoder.}
In our Transformer decoder, we use autoregressive method to predict one fixation each time and iteratively generate the whole fixation sequence.
Through our observation, more and more models try to adopt a parallel structure in the Transformer decoder\cite{Detr,MDETR,GazeFormer} for higher efficiency. Therefore, we also tried to build a parallel Transformer decoder in our method (denoted as `Parallel'). 
In the parallel model, we use a random embedding with length $L$ as the input to the Transformer decoder, and we can get the hidden state of a scanpath of length L at one time, all the other parts of the model are unchanged. 

We can observe from \cref{Ablation_table} that the parallel structure of the decoder performs poorly than the autoregressive Transformer decoder, especially in the TDE\cite{TDE} metric. Such comparison results demonstrate that an autoregressive Transformer decoder is more suitable for our scanpath prediction task.

\section{Limitaions and Future Works}

In our work, we employed SphereNet\cite{S-CNN} for feature extraction and 3D Mixture Density Network (MDN) for fixation generation. However, we acknowledge that these components are replaceable, and we are contemplating whether there might be better alternatives for both parts.

In addition, we are considering incorporating relevant semantic information to improve the performance further, as objects with important semantic information tend to attract human attention more easily.

\section{Conclusion}
Considering that the 2D equirectangular projection of 360° images has the issues of distortion and coordinate discontinuity, we proposed to execute scanpath prediction for 360° images in 3D spherical coordinate system. Correspondingly, we proposed a novel 3D Scanpath Transformer named Pathformer3D, where a 3D Transformer Encoder was first used to obtain 3D contextual feature representation of the 360° image. Then, Pathformer3D input the 3D contextual feature representation and historical fixation information into the Transformer Decoder to predict the hidden state for each fixation. Each fixation's hidden state was then fed into a 3D mixture density network to output the 3D Gaussian distribution of the fixation, from which the fixation location can be sampled. 
Finally, Pathformer3D model was compared with three state-of-the-art 360° image scanpath prediction models, to demonstrate the superiority of its performance. 

\subsubsection{Acknowledgements}
This work is supported in part by the National Natural Science Foundation 
of China under grant 62206127 and 62272229, the Natural Science Foundation of Jiangsu Province under grant BK20222012, and Shenzhen Science and Technology Program JCYJ20230807142001004. Rong Quan and Yantao Lai contributed equally to this work. Dong Liang is the corresponding author.


%
%
\bibliographystyle{splncs04}
\bibliography{main}
\end{document}